\documentclass[10pt,twocolumn,letterpaper]{article}

\usepackage{cvpr}              

\usepackage[dvipsnames]{xcolor}

\definecolor{cvprblue}{rgb}{0.21,0.49,0.74}
\usepackage[pagebackref,breaklinks,colorlinks,citecolor=cvprblue]{hyperref}
\usepackage{multirow}
\def\paperID{*****} 
\def\confName{CVPR}
\def\confYear{2024}

\title{PCIE\_LAM Solution for Ego4D Looking At Me Challenge}

\author{
Kanokphan Lertniphonphan \\
Lenovo Research\\
{\tt\small klertniphonp@lenovo.com}
\and
Jun Xie \\
Lenovo Research\\
{\tt\small xiejun@lenovo.com}
\and
Yaqing Meng \\
Chinese Academy of Sciences\\
{\tt\small mengyaqing2023@ia.ac.cn}
\and
Shijing Wang \\
Beijing Jiaotong University\\
{\tt\small shijingwang@bjtu.edu.cn}
\and
Feng Chen\\
Lenovo Research\\
{\tt\small chenfeng13@lenovo.com}
\and
Zhepeng Wang \\
Lenovo Research\\
{\tt\small wangzpb@lenovo.com}
}

\begin{document}
\maketitle
\begin{abstract}

This report presents our team's 'PCIE\_LAM' solution for the Ego4D Looking At Me Challenge at CVPR2024. The main goal of the challenge is to accurately determine if a person in the scene is looking at the camera wearer, based on a video where the faces of social partners have been localized. Our proposed solution, InternLSTM, consists of an InternVL image encoder and a Bi-LSTM network. The InternVL extracts spatial features, while the Bi-LSTM extracts temporal features. However, this task is highly challenging due to the distance between the person in the scene and the camera movement, which results in significant blurring in the face image. To address the complexity of the task, we implemented a Gaze Smoothing filter to eliminate noise or spikes from the output. Our approach achieved the 1$^{st}$ position in the looking at me challenge with 0.81 mAP and 0.93 accuracy rate. Code is available at \url{https://github.com/KanokphanL/Ego4D_LAM_InternLSTM}

\end{abstract}

\section{Introduction}
\label{sec:intro}

Recently, an egocentric video captured using a wearable camera has gained significant importance in the fields of computer vision and robotics. In the Ego4D social benchmark, the interaction between the camera wearer and the participant is highlighted in the Looking At Me (LAM) challenge \cite{Grauman2021Ego4DAT}. 

Various approaches \cite{Grauman2021Ego4DAT, gazepose} have been proposed to accurately analyze the interaction of the participant with the camera wearer in image sequences. In \cite{gazepose}, landmarks and head pose estimation techniques were employed to improve spatial feature refinement. Subsequently, the spatial feature refinement and temporal dynamic refinement were applied on \cite{Grauman2021Ego4DAT}. 

In this work, we introduce InternLSTM, a method for extracting spatial and temporal features from sequences of images to enhance performance. Furthermore, we apply a smoothing filter in the post-processing stage to eliminate spikes caused by motion blur and low quality face image.

\section{Method}
\label{sec:method}

\subsection{InternLSTM}

\begin{figure}[t]
  \centering
   \includegraphics[width=1\linewidth]{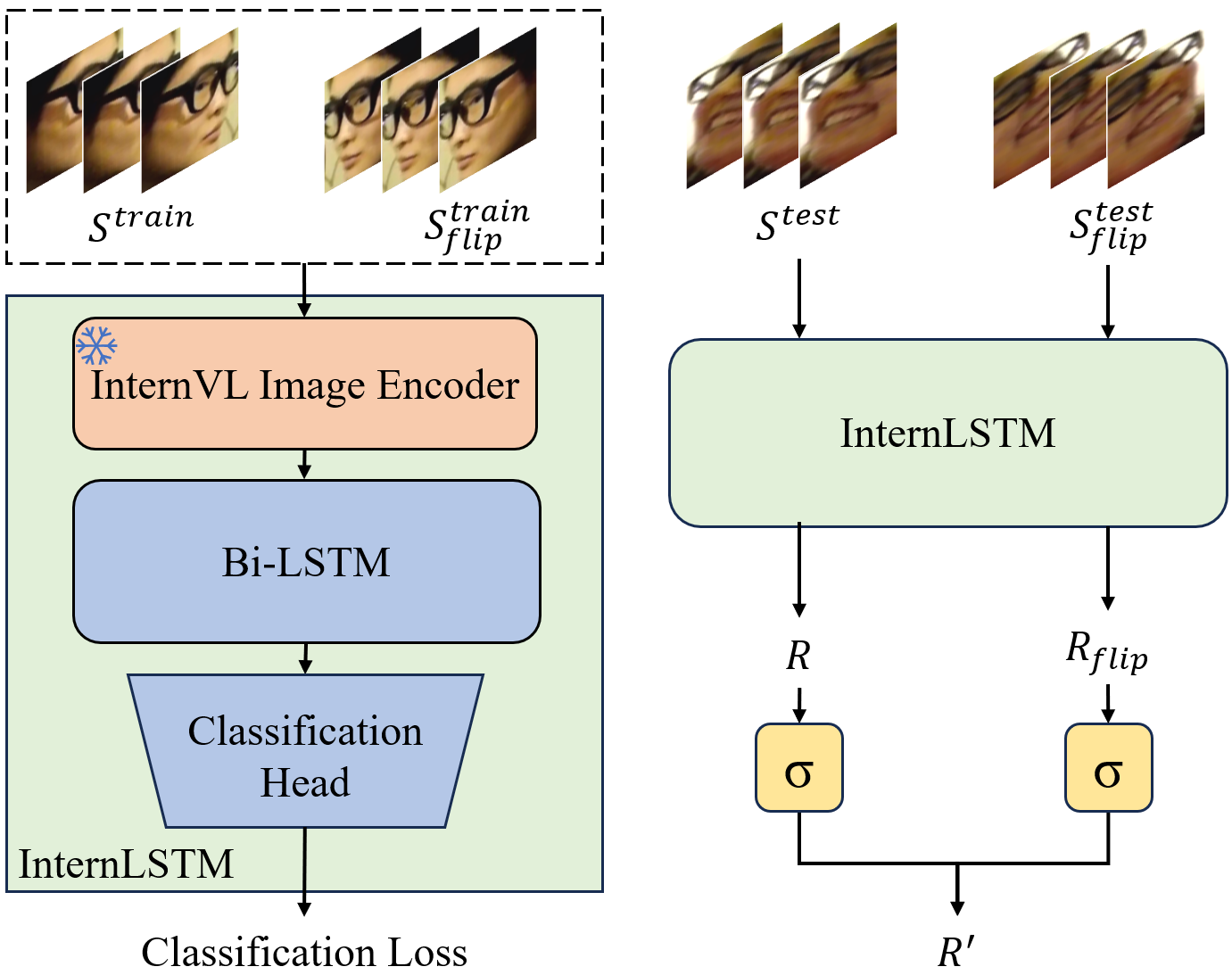}
   \caption{
   The Training and Inference Framework for InternLSTM.
   }
   \label{fig:InternLSTM}
\end{figure}

Using the strong visual representation capabilities of large language models, we introduce InternLSTM based on InternVL\cite{chen2023internvl}. It consists of the InternVL image encoder, Bi-LSTM, and a classification head as shown in figure \ref{fig:InternLSTM}. The InternVL image encoder was freezed during training, while all other modules undergo training. 

Initially, the spatial features of the image encoder are down-sampled. Then, the Bi-LSTM network extracts temporal features, capturing intermediary frame features. A classification head with 4 fully connected layers is utilized along with the cross entropy loss function. 

\begin{table*}
  \centering
  \begin{tabular*}{0.8\textwidth}{@{\extracolsep{\fill}}lcccc} 
    \toprule
     &  \multicolumn{2}{c}{Validation}  &  \multicolumn{2}{c}{Test} \\
    \hline
    \multicolumn{1}{c}{Method}  & mAP ($\uparrow$) & Acc ($\uparrow$) & mAP ($\uparrow$) & Acc ($\uparrow$)  \\
    \midrule
    InternLSTM & 84.23\% & 94.79\% & 76.57\% & 90.00\%\\
     + Horizontal flip when training & 83.66\% & 91.93\% & 76.94\% & 80.04\%\\
     + TTA & - & - & 77.18\% & 80.30\%\\
    \bottomrule
  \end{tabular*}
  \caption{InternLSTM performance on different training and testing setting.}
  \label{tab:intern_test}
\end{table*}

\begin{table*}
    \centering
    \begin{tabular*}{0.8\textwidth}{@{\extracolsep{\fill}}lcccc} 
    \toprule
         &  \multicolumn{2}{c}{Validation}  &  \multicolumn{2}{c}{Test} \\
        \hline
        \multicolumn{1}{c}{Method}  & mAP ($\uparrow$) & Acc ($\uparrow$) & mAP ($\uparrow$) & Acc ($\uparrow$)  \\
        \midrule
        GazeLSTM \cite{Grauman2021Ego4DAT} & 79.40\% & 92.78\% & 74.54\% & 87.04\% \\ 
        GazePose \cite{gazepose} & 81.89\% & 91.84\% & 76.95\% & 85.08\% \\
        GazePose with Gaze Smoothing & 82.69\% & 92.44\% & 78.50\% & 85.92\%\\
        \bottomrule
  \end{tabular*}
  \caption{Baseline and GazePose performance}
  \label{tab:base}
\end{table*}

Meanwhile, since reversing images should not impact the prediction output, we augment the original training dataset $S^{train}$ by horizontal flip to create $S^{train}_{flip}$. Both $S^{train}$ and $S^{train}_{flip}$ are used in the training. When testing on the test dataset, we also flip the original test dataset.  Applying InternLSTM to the original test dataset $S^{test}$ and the flipped test dataset $S^{test}_{flip}$, we obtain $R$ and $R_{flip}$, respectively. Subsequently, softmax is used to obtain the final score.

\subsection{Test-Time Augmentation (TTA) and Ensemble}

During the validation and testing, we implemented TTA to enhance performance. Specifically, only the horizontal flip technique was utilized in TTA for this LAM dataset.

In addition to InternLSTM, we reproduced the baseline results \cite{Grauman2021Ego4DAT} and GazePose \cite{gazepose}. To combine these results, ensemble methods  \cite{roman2021weight} were utilized. The overall ensemble score was computed as the mean of the combined prediction scores.

\subsection{Gaze Smoothing}

In this dataset, images from the wearable camera contain significant blurring in the face image. The image with motion blur reduces the performance of gaze direction or head pose estimation from the face image. Moreover, transitions in behavior take some time compared to sudden changes induced by motion noise. To address this, a median smoothing filter was applied as a post-processing to mitigate spikes in prediction scores. This filter calculates the median value within a specified time window and replaces the central score with the median value.

\section{Experiments}
\label{sec:experiment}

\subsection{Dataset and Evaluation metric}

We conducted experiments using the Ego4D social benchmarks \cite{Grauman2021Ego4DAT}. The dataset comprises 572 clips, each lasting 5 minutes. 389 and 50 clips are allocated for training and validation, respectively, with the remainder set aside for testing. Ground truth includes a binary label for each face in every frame within the time segment showing a person looking at the camera wearer. Mean Average Precision (mAP) and Top-1 accuracy are used for evaluation.

\subsection{Experimental Setting} 

First, we employed InternVL on 4 A100 (80GB) GPUs to extract features from all images, including both original and flipped version. Then, InternLSTM was trained with a batch size of 128 using the Adam optimizer and a learning rate of 1e-4.

\begin{figure}[b]
  \centering
   \includegraphics[width=0.6\linewidth]{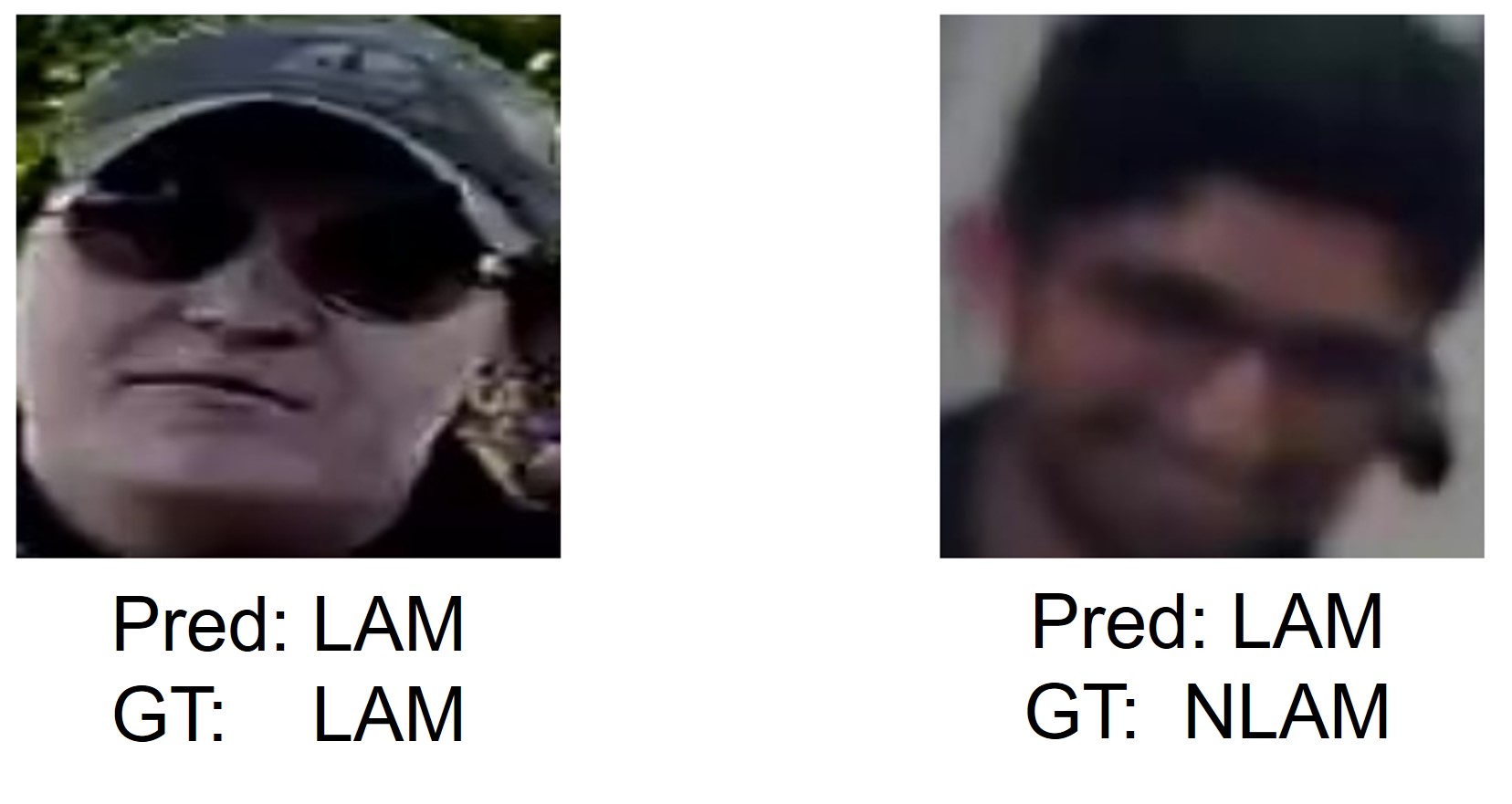}
   \caption{
   The InternLSTM performance varies with different image qualities. The model incorrectly predicted labels for blurry images as shown in the right image.   
   }
   \label{fig:intern_lim}
\end{figure}

\subsection{Results}

The results from InternLSTM are presented in table \ref{tab:intern_test}. We experimented InternLSTM using original training images and flipped augmentation. Our InternLSTM model was initially trained without any augmentation and achieved 84.23\% mAP and 94.79\% accurancy on the validation set. Then we train the model with horizontal flip and achieved 83.66\% mAP and 91.93\% accuracy. Finally, TTA (horizontal flip only) was applied during inference and achieve the highest score in test set. 

InternLSTM performs well in predicting results when images are clear. However, its predictions deviate from the ground truth when the image quality is low, as illustrated in figure \ref{fig:intern_lim}.

We also reproduced the baseline \cite{Grauman2021Ego4DAT} and GazePose \cite{gazepose} approaches as shown in table \ref{tab:base}. We added Gaze smoothing to GazePose results and improved the performance in both validation and testing.

From the InternLSTM results, data augmentation and TTA improved the testing results, but the validation results did not show the same improvement for our model. Comparing the mAP in table \ref{tab:intern_test} and \ref{tab:base}, a higher mAP in validation may not always result in better testing performance. This issue also occurred with other approaches we attempted, making model selection challenging.

Our final results combined outputs from several approaches including InternLSTM, GazeLSTM, and GazePose with various parameter configurations. The results after applying gaze smoothing are presented in table \ref{tab:final}. Our approach achieved the 1$^{st}$ position in the Ego4D Looking At Me leaderboard with an mAP of 0.81 and an accuracy of 0.93.


\begin{table}
  \centering
  \begin{tabular}{l c c}
    \toprule
    \multicolumn{1}{c}{Team} & mAP ($\uparrow$) & Acc ($\uparrow$)  \\
    \midrule
    PCIE\_LAM (Ours) & \textbf{0.81} & \textbf{0.93} \\
    listen & 0.80 & 0.87 \\
    PKU-WICT-MIPL & 0.79 & 0.92 \\
    ydejie & 0.79 & 0.92 \\
    Dejie (gb6) & 0.79 & 0.92 \\
    Host\_24191\_Team & 0.66 & 0.74 \\ 
    \bottomrule
  \end{tabular}
  \caption{Ego4D Looking at me Challenge Leaderboard}
  \label{tab:final}
\end{table}


\section{Conclusion}

We introduce InternLSTM consists of an InternVL image encoder and a Bi-LSTM for spatial and temporal feature extraction from sequence images. The data augmentation and TTA helped improve model performance. We also applied ensemble and gaze smoothing to increase accuracy and reduce noise from the outputs. Our next step is to enhance robustness of the model over the blur input images.

{
    \small
    \bibliographystyle{ieeenat_fullname}
    \bibliography{main}
}


\end{document}